\documentclass[pmlr]{jmlr}

\usepackage{color}
\usepackage{amsmath}
\usepackage{bm}
\usepackage{url}
\usepackage{multirow}
\usepackage{threeparttable}
\usepackage{algorithm}
\usepackage{algorithmic}

\jmlrvolume{85}
\jmlryear{2018}
\jmlrworkshop{Machine Learning for Healthcare}

\title[Predicting Smoking with a Hawkes Process Model]{Predicting Smoking Events with a Time-Varying Semi-Parametric Hawkes Process Model}

\author{\Name{Matthew Engelhard}\thanks{These authors contributed equally to this work} \Email{m.engelhard@duke.edu} \\
	\addr Department of Psychiatry and Behavioral Sciences\\
	Duke University School of Medicine\\
	Durham, NC, USA
	\AND
	\Name{Hongteng Xu}\footnotemark[1] \Email{hongteng.xu@duke.edu} \\
	\addr Department of Electrical and Computer Engineering\\
	Duke University\\
	InfiniaML, Inc.\\
	Durham, NC, USA
	\AND
	\Name{Lawrence Carin} \Email{lcarin@duke.edu} \\
	\addr Department of Electrical and Computer Engineering\\
	Duke University\\
	Durham, NC, USA
	\AND
	\Name{Jason A Oliver} \Email{jason.a.oliver@duke.edu} \\
	\addr Department of Psychiatry and Behavioral Sciences\\
	Duke University School of Medicine\\
	Durham, NC, USA
	\AND
	\Name{Matthew Hallyburton} \Email{matt.hallyburton@duke.edu} \\
	\addr Department of Psychiatry and Behavioral Sciences\\
	Duke University School of Medicine\\
	Durham, NC, USA
	\AND
	\Name{F Joseph McClernon} \Email{francis.mcclernon@duke.edu} \\
	\addr Department of Psychiatry and Behavioral Sciences\\
	Duke University School of Medicine\\
	Durham, NC, USA}

\editor{}

\begin{document}

\maketitle

\begin{abstract}
  Health risks from cigarette smoking --- the leading cause of preventable death in the United States --- can be substantially reduced by quitting. Although most smokers are motivated to quit, the majority of quit attempts fail. A number of studies have explored the role of self-reported symptoms, physiologic measurements, and environmental context on smoking risk, but less work has focused on the temporal dynamics of smoking events, including daily patterns and related nicotine effects. In this work, we examine these dynamics and improve risk prediction by modeling smoking as a self-triggering process, in which previous smoking events modify current risk. Specifically, we fit smoking events self-reported by 42 smokers to a time-varying semi-parametric Hawkes process (TV-SPHP) developed for this purpose. Results show that the TV-SPHP achieves superior prediction performance compared to related and existing models, with the incorporation of time-varying predictors having greatest benefit over longer prediction windows. Moreover, the impact function illustrates previously unknown temporal dynamics of smoking, with possible connections to nicotine metabolism to be explored in future work through a randomized study design. By more effectively predicting smoking events and exploring a self-triggering component of smoking risk, this work supports development of novel or improved cessation interventions that aim to reduce death from smoking.
  
\end{abstract}

\section{Introduction}

Cigarette smoking is responsible for over 480,000 deaths per year in the United States, with total yearly economic cost exceeding \$300 billion \citep{CDC2014}. Quitting smoking quickly and substantially lowers the risk of heart attack, stroke, and cancer \citep{CDC2010}, and most adult smokers are motivated to quit \citep{Jamal2018}. Unfortunately, fewer than one in three smokers who attempt to quit use evidence-based methods --- often due to barriers to access --- and there is a large divide between yearly quit attempts ($>$50\%) and successful quitting ($<$10\%) \citep{Babb2017}. Consequently there is an ongoing and urgent need to develop novel, effective, and accessible cessation interventions.

To more effectively support cessation, we can seek to better understand the immediate causes (or precursors) of smoking events. Currently, there is limited understanding of the factors leading to lapse (i.e. smoking during a quit attempt), and the process through which a single lapse becomes a relapse (i.e. failed quit attempt). If moments of high lapse risk can be identified, possible interventions include accessing self-help materials \citep{Siu2015}, telephone support/counseling \citep{Siu2015,Stead2013}, or use of short-acting pharmacologic treatment such as a nicotine lozenge \citep{Kotlyar2017}, oxytocin \citep{Miller2016}, or a nicotine film \citep{Du2015}. Smoking risk has been estimated based on passive physiologic measurements \citep{Chatterjee2016}, self-reported symptoms (e.g. urge, stress) \citep{Businelle2016}, and previously observed daily patterns \citep{Chandra2011}. Studies in controlled settings have also suggested that the surrounding environment plays an important role \citep{McClernon2016,Stevenson2017}. However, less emphasis has been placed on the underlying temporal dynamics of smoking events --- in other words, the natural rhythms and patterns of smoking within and between days. The goal of this work is to improve prediction and knowledge of smoking risk by utilizing a prediction model that can learn these dynamics.



Temporal patterns of smoking have been studied since the 1970s, when \citet{Frederiksen1977} and \citet{Robinson1980} independently observed consistent daily patterns of smoking in several small cohorts of smokers. These patterns were even more striking when considering individual participants, whose behaviors were more consistent than the cohorts as a whole. More recent work by \citet{Chandra2007} has strengthened and extended this line of research with much larger study (n=351) that identified four common daily smoking patterns (e.g. daily decline, morning high). Subsequently these patterns were confirmed and additionally tied to craving, including evidence that craving tends to precede smoking \citep{Chandra2011}. The connection between smoking patterns and craving --- a factor which also predicts failure to quit \citep{Killen1997} --- is strong motivation for predicting craving \citep{Chatterjee2016} as a proxy for smoking risk to inform an intervention. Nevertheless risk and craving are distinct, and their relative merits in predicting smoking and/or lapses are currently unknown.

Related to daily patterns, smoking is influenced by plasma and saliva nicotine concentrations, which peak within minutes of smoking and return to baseline over several hours ($\sim$2hr plasma half-life) \citep{Benowitz2009}. Plasma nicotine concentration and craving are negatively correlated, suggesting that craving is driven in part by nicotine seeking \citep{Jarvik2000}. \citet{Gomeni2001} directly studied the relative effects of nicotine and circadian patterns on craving with a blinded study design, finding that nicotine replacement can attenuate craving. Indeed, this effect is the motivation for nicotine-based cessation treatment, which has been shown to be effective \citep{Kotlyar2017,Du2015}. Given the strong relationship between craving and smoking itself \citep{Chandra2011}, these results suggest that cigarette smoking is a self-triggering (i.e. self-reinforcing) process in which the probability of smoking is modulated by previous smoking events.

To our knowledge, this self-triggering dynamic has not yet been incorporated in smoking prediction models. For example, \citet{rathbun2016mixed} model smoking events as a Poisson process with intensity function (i.e. rate of smoking) modulated by self-reports of affect, arousal, attention, and restlessness; but in their model, previously observed smoking events do not directly modify the current intensity. Here we propose to explore the self-triggering nature of smoking with a Hawkes process, wherein the intensity is influenced not only by exogenous factors --- such as demographics or self-reports --- but also the accumulating influence of previous smoking events.

Specifically, we develop the time-varying semi-parametric Hawkes process (TV-SPHP), then fit it to smoking events reported by smokers (n=42) over a $>$1 week period. This approach allows us to capture the effects of the following predictors on current smoking intensity (i.e. risk): (a) fixed covariates such as demographics; (b) time-varying covariates, including speed as measured by GPS; and (c) previous smoking events. The relationship between recent smoking and current risk is modeled with a semi-parametric \textit{impact function}, which is itself a sum of parametric basis functions.

From a clinical perspective, this approach offers two important advantages. First, it allows us to investigate the self-triggering nature of smoking in the real world by examining the learned impact function. We hypothesize that the impact function will exhibit daily periodicity and peak within 1-4 hours of smoking, which may reflect increased nicotine seeking. Second, it may improve prediction performance compared to alternative approaches by modeling the effect of previous events on current risk. We hypothesize that the TV-SPHP will improve prediction performance --- measured as the mean absolute error over periods ranging from 10 minutes to 1 hour --- compared to baseline and existing models.

\paragraph{Technical Significance}
This work develops the time-varying, semi-parametric Hawkes process (TV-SPHP) and its associated learning/inference algorithm. The TV-SPHP extends previous Hawkes process variants by modeling the intensity function as a sum of exogenous and endogenous factors. The exogenous factors include both time-invariant and time-varying participant features, while the endogenous factors capture the self-triggering nature of smoking using a semi-parametric impact function.

\paragraph{Clinical Relevance}
The clinical significance of this work is twofold. First, we improve on existing smoking prediction models and establish a new performance baseline. Smoking prediction models can identify high risk contexts either (a) in real-time to trigger a just-in-time adaptive intervention, or (b) in advance to help smokers avoid these contexts. Second, we connect the self-triggering nature of smoking to nicotine metabolism as a hypothesis to be explored further in future work.

\section{Related Work}

\subsection{Smoking Prediction}



Compared to the smoking detection literature --- which is intimately connected to the large, rapidly progressing activity recognition literature \citep{ordonez2016deep} --- less work has focused on smoking prediction. \citet{Dumortier2016} compared the performance of several machine learning classifiers (Naive Bayes, discriminant analysis, decision tree) to discriminate between high-urge and low-urge states based on self-reported contextual information. \citet{Chatterjee2016} also predicted high and low craving with an approach based on conditional random fields, but their system was based on physiologic measurements passively collected with wearable devices.

The work of \citet{rathbun2016mixed} is most closely related to the current work. A total of 304 smokers self-reported all smoking events along with contextual information with a hand-held device. These events were fitted to an inhomogeneous Poisson process with intensity modified by the time-varying contextual predictors. In contrast to our work, (a) the focus was model interpretation and not prediction; (b) their model did not include a self-triggering component; and (c) time-varying covariates were self-reported contextual information rather than GPS measurements.

\subsection{Temporal point processes and Hawkes processes}

Temporal point processes~\citep{daley2007introduction} have been widely used to model real-world event sequences. 
In the simplest 1D case, given an individual's sequence of events $\bm{s}=\{t_1,t_2,...|t_i\in[0,T)\}$ and the corresponding counting process $N(t)=|\{t_i\in\bm{s}|t_i\leq t \}|$, a point process model characterizes the sequence by modeling the expected instantaneous event occurrence rate over time:
\begin{eqnarray}\label{intensity}
\begin{aligned}
\lambda(t) = \frac{\mathbb{E}[dN(t)|\mathcal{H}_t]}{dt},
\end{aligned}
\end{eqnarray}
where $\mathcal{H}_t=\{t_i\in\bm{s}|t_i\leq t \}$ contains all historical events happening before or at time $t$ and $\mathbb{E}[dN(t)|\mathcal{H}_t]$ computes the expected number of events in $[t, t+dt)$ conditioned on the history. 
Generally, we call $\lambda(t)$ the \emph{intensity function}. 
The key to the point process model is modeling and learning $\lambda(t)$ based on observed data.

A important kind of point process called the Hawkes process~\citep{hawkes1971point} is useful for explicitly modeling sequences whose events are self- and mutually-triggering. Mathematically, the intensity function of a 1D Hawkes process has the following form:
\begin{eqnarray}\label{hp}
\begin{aligned}
\lambda(t)= \mu(t) + \int_{0}^{T}\phi(t-s)dN(s)=\mu(t) + \sideset{}{_{t_i\in\mathcal{H}_t}}\sum\phi(t-t_i).
\end{aligned}
\end{eqnarray}
where $\mu(t)$ corresponds to the background intensity caused by exogenous factors~\citep{xu2017benefits}, which can be modeled as homogeneous or inhomogeneous Poisson process. 
The term $\int_{0}^{T}\phi(t-s)dN(s)$ represents the accumulation of endogenous intensity caused by the history~\citep{farajtabar2014shaping}. 
The nonnegative function $\phi(t-s)$, where $t\geq s$, is called the \emph{impact function}. It represents the influence of an historical event at time $s$ on the intensity at time $t$. 

Recently, Hawkes processes have attracted many researchers, and a number of variants have been proposed. These include the mixture of Hawkes processes~\citep{xu2017dirichlet}, the nonlinear Hawkes process~\citep{xu2017patient}, and the locally-stationary Hawkes process~\citep{roueff2016locally,xu17b}. 
Applications include network analysis~\citep{zhao2015seismic}, quantitative finance~\citep{da2014hawkes,bacry2015hawkes}, and e-health~\citep{xu2017patient}. 
Maximum likelihood estimation (MLE) is one of the most popular approaches for learning Hawkes processes~\citep{lewis2011nonparametric,zhou2013learning}. 
In addition to MLE, least squares methods~\citep{eichler2017graphical}, Wiener-Hopf equations~\citep{bacry2015hawkes}, and the Cumulants-based method~\citep{achab2016uncovering} have been applied. 
However, stochastic optimization for Hawkes processes has not been studied systematically. 
For nonparametric Hawkes processes, online learning methods were proposed in~\citep{hall2016tracking,yang2017online}, but they use time-consuming discretization or kernel-estimation when learning models, and thus have poor scalability. 
The risk bound and the sample complexity of learning a single Hawkes process are investigated in~\citep{daneshmand2014estimating,eichler2017graphical,yang2017online}.

\section{Cohort}

\subsection{Participants and Study Design}

Adult smokers (N=60) were recruited from the Durham, NC area. All participants completed an IRB-approved informed consent form. Participant characteristics (age, sex, race, ethnicity, marital status, highest education completed, number of years smoked, and cigarettes smoked per day) were collected during a baseline study visit. After the visit, smokers carried a small, rechargeable GPS tracker for approximately one week. The trackers are smaller than a cigarette pack and fit easily in a pocket or purse. Participants were required to keep the tracker on their person at all times except when it could become wet (e.g. while bathing or swimming), and they received detailed instructions on tracker use and recharging. 

A button on the front of the tracker can be pressed to indicate the times of events of interest. Participants were instructed to manually log all cigarettes smoked by pressing this button just before lighting the cigarette. \citet{shiffman2002immediate} showed that smoking events self-reported with a button press were highly correlated with cotinine and exhaled CO levels as well as participant recall on timeline follow-back assessment. Nevertheless, we took additional steps (see Section 3.3) to validate these data.



\subsection{Inclusion and exclusion criteria}

Participants were required to be generally healthy and between the ages of 18 and 55. They were also required to have smoked at least 5 cigarettes with normal nicotine content ($>$0.5mg) per day for at least a year, with an afternoon expired CO concentration $>$8 ppm.

Those with current or previous significant health problems (e.g. chronic hypertension, COPD, seizure disorder, liver or kidney disorder, coronary heart disease, myocardial infarction, arrhythmia) were excluded. Other exclusion criteria included use of psychoactive medications, use of smokeless tobacco, current alcohol or drug abuse (confirmed by urine drug screen), and use of nicotine replacement therapy or other smoking cessation treatment.

\subsection{Data collection and pre-processing}

Raw data downloaded from the GPS trackers included GPS coordinates, GPS performance level (coordinate accuracy), estimated speed, the time of measurement, and a button press indicator. These data were collected at 30 second intervals throughout the collection period with two exceptions. First, the trackers incorporate a ``smart tracking" functionality that prevents GPS collection when no movement is detected to conserve battery life. Second, each logged smoking event (button press) prompted an additional measurement at that time. Available features are summarized in Table 1.

\begin{table}[t]
  \centering 
  \small{
  \caption{The list of collected features}\label{tab1}
  \begin{tabular}{l|l}\hline\hline
    {Feature Domain} & 
    {Options} \\
    \hline
    Exhaled CO (ppm)&1. $<10$; 2. $[10, 20]$; 3. $[20, 30]$; 4. $>30$\\
    \# Years Smoked &1. $<10$; 2. $[10, 20]$; 3. $>20$\\
    Age &1. $<25$; 2. $[20, 25]$; 3. $>50$\\
    Sex &1. Male; 2. Female\\
    Race &1. White; 2. American India; 3. Asian; 4. Black; 5. Hispanic; 6. Other\\
    Ethnicity &1. Non-Hispanic; 2. Hispanic; 3. Unknown\\
    Education &1. Elementary; 2. Middle; 3. High; 4. Coll/Tech; 5. College Graduate\\
    Marital Status &1. Single; 2. Separated; 3. Widowed; 4. Married; 5. Divorced\\
    GPS Measurements &Longitude; Latitude; Speed (km/h); Distance (m)\\
    Time of Day &1. Night; 2. Morning; 3. Afternoon; 4. Evening
    \\ \hline\hline 
  \end{tabular}
  \label{tab:example} 
  }
\end{table}

The raw data from all N=60 participants were first pre-processed to exclude participants for whom there was strong evidence of unreliable button presses. This was determined by partitioning each participant's data into individual days and comparing (a) the rate of daily button presses to the expected daily rate (based on self-reported cigarettes per day) by nonparametric two-sample proportion test; and (b) comparing the rate of button presses over the entire collection period to the expected rate. Participants were excluded when $p<0.01$ for two or more days --- an event expected to occur with approximately 0.002 probability given 7 days of collection (21 pairs of days) --- or when $p<0.01$ for the collection period as a whole.

As a final pre-processing step, repeated button presses ($<$60 seconds apart) were removed. Multiple rapid sequences of presses were observed in several participants. Since the tracker does not provide feedback/confirmation when the button is pressed, these sequences may reflect the participant's uncertainty as to whether their button presses have been recorded. The 60 second threshold was chosen was a reasonable lower bound on the time between successive cigarettes. We assume that these sequences do represent a smoking event, however, so the final button press in each such sequence was retained.

\section{Proposed method}
\subsection{Time-varying Semi-parametric Hawkes processes}
Consider $N$ individuals' smoking behaviors, $i.e.$, $\mathcal{S}=\{\bm{s}_{n}\}_{n=1}^{N}$, where $\bm{s}_n=\{t_1^n, t_2^n,...\}$ records the timestamps of the $n$-th individual's smoking behaviors.
Participant characteristics ($e.g.$, race, gender, age and education) are recorded as a time-invariant feature $\bm{f}_0^n\in\mathbb{R}^C$. 
A participant's motion status ($e.g.$, location and speed) is also recorded as a time-varying feature vector $\bm{f}^n(t)\in\mathbb{R}^D$.

For each individual, the sequence of his/her smoking behaviors is an instantiation of a temporal point process system. 
As previously mentioned, the dynamics of the system ($i.e.$, the occurrence of a smoking behavior at certain time) are influenced by three factors: an intrinsic propensity of that individual, the current environment and status, and the triggering from historical smoking events. 
From the viewpoint of Hawkes processes, the first two factors are history-independent, and may be viewed as exogenous factors of the system; while the remaining factor depends on historical behaviors, and is the endogenous factor of the system. 
As this analysis suggests, the Hawkes process is well-suited for capturing factors affecting smoking behavior in its intensity function.
In particular, we propose the following semi-parametric Hawkes process model for smoking behaviors. 
For each individual $n$, the intensity function is written as:
\begin{eqnarray}\label{sphp}
\begin{aligned}
\lambda^n(t) = \underbrace{\bm{\mu}_{TI}^{\top}\bm{f}_0^n +\bm{\mu}_{TV}^{\top}\bm{f}^n(t)}_{\text{exogenous factors}}+\underbrace{\sideset{}{_{t_i^n<t}}\sum \phi(t-t_i)}_{\text{endogenous factor}},~\text{for}~n=1,...,N,
\end{aligned}
\end{eqnarray}
where $\bm{\mu}_{TI}\in\mathbb{R}^C$ and $\bm{\mu}_{TV}\in\mathbb{R}^D$ are parameters corresponding to time-invariant and time-varying features, respectively.
The $\bm{\mu}_{TI}$, $\bm{\mu}_{TV}$, and impact function $\phi(t)$ are the parameters of our model, which are shared across all subjects. 

In contrast to traditional parametric or nonparametric Hawkes process models~\citep{zhou2013learning,zhou2013learning2,eichler2017graphical}, here we design a time-varying semi-parametric Hawkes process (TV-SPHP) model: the exogenous intensity $\mu(t)$ in (\ref{hp}) is parametrized as $\bm{\mu}_{TI}^{\top}\bm{f}_0^n +\bm{\mu}_{TV}^{\top}\bm{f}^n(t)$ to take advantage of all available features, while the impact function $\phi(t)$ is modeled nonparametrically to reduce the risk of model misspecification.
In particular, we use the basis representation method to model the impact function as
\begin{eqnarray}\label{impact}
\begin{aligned}
\phi(t) = \sideset{}{_{m=1}^{M}}\sum a_m\kappa_m(t),
\end{aligned}
\end{eqnarray}
where $\{\kappa_m(t)\}_{m=1}^{M}$ are predefined basis functions ($e.g.$, wavelets, Gabor, Gaussian kernel, etc.) and $\bm{a}=\{a_m\}_{m=1}^M$ are the corresponding coefficients. 
Given our training event sequences, we apply the basis selection method proposed in~\citep{xu2016learning} to decide the number and bandwidth of the basis functions.

\subsection{Learning algorithm}
Given training event sequences $\mathcal{S}=\{\bm{s}^n\}$ from $N$ subjects and the proposed model mentioned above, we can learn the parameters of the model via maximum likelihood estimation (MLE), which is a common method used in many existing works~\citep{lewis2011nonparametric,zhou2013learning,luo2015multi}. 
In particular, the learning problem of our model is the following optimization problem:
\begin{eqnarray}\label{loss}
\begin{aligned}
\sideset{}{_{\bm{\theta}\geq\bm{0}}}\min\mathcal{L}(\bm{\theta}; \mathcal{S})+\gamma\mathcal{R}(\bm{\theta}),
\end{aligned}
\end{eqnarray}
where $\bm{\theta}=[\bm{\mu}_{TI};\bm{\mu}_{TV};\bm{a}=[a_m]]\in\mathbb{R}^{C+D+M}$ are the parameters of our model.
According to~\citet{daley2007introduction}, the negative log-likelihood function $\mathcal{L}(\bm{\theta};\mathcal{S})$ is
\begin{eqnarray}\label{nll}
\begin{aligned}
\mathcal{L}(\bm{\theta};\mathcal{S})=\sideset{}{_{n=1}^{N}}\sum\Bigl\{ \int_{0}^{T}\lambda^n(s)ds - \sideset{}{_{i=1}^{I_n}}\sum \log\lambda^n(t_i^n) \Bigr\},
\end{aligned}
\end{eqnarray}
where $T$ is the length of training event sequence, and $I_n$ is the number of events of the $n$-th sequence. 
Based on (\ref{sphp}) and (\ref{impact}), it is easy to prove that (\ref{nll}) is convex~\citep{zhou2013learning}. 
$\mathcal{R}(\bm{\theta})$ is an arbitrary convex regularizer and its significance is controlled by the weight $\gamma$. 
In this work, we only impose sparsity on the parameters, $i.e.$, $\mathcal{R}(\bm{\theta})=\|\bm{\theta}\|_1=\|\bm{\mu}_{TI}\|_1+\|\bm{\mu}_{TV}\|_1+\|\bm{a}\|_1$, where $\|\cdot\|_1$ denotes the $\ell_1$ norm. 

Similar to the algorithms in~\citep{zhou2013learning}, we introduce an auxiliary variable and a dual one for $\bm{\theta}$, denoted as $\bm{y}$ and $\bm{z}$, and rewrite (\ref{loss}) as a Lagrangian form:
\begin{eqnarray}\label{lagrange}
\begin{aligned}
\sideset{}{_{\bm{\theta},\bm{y},\bm{z}\geq\bm{0}}}\min \widetilde{\mathcal{L}}(\bm{\theta},\bm{z},\bm{y};\mathcal{S})=\mathcal{L}(\bm{\theta}; \mathcal{S})+\gamma\|\bm{z}\|_1 + \rho\bm{y}^{\top}(\bm{\theta}-\bm{z}) + \frac{\rho}{2}\|\bm{\theta}-\bm{z}\|_2^2.
\end{aligned}
\end{eqnarray}
This problem can be solved iteratively by the alternating direction method of multipliers (ADMM)~\citep{zhou2013learning}. 
In particular, the scheme of our learning algorithm is shown in Algorithm~\ref{alg1}, where the function $(\cdot)_{+}$ in line 5 sets all negative elements to zeros and the function $S_{\tau}(\cdot)$ in line 6 is the soft-thresholding function with threshold $\tau$.  
The derivation details of the gradient of $\mathcal{L}(\bm{\theta};\mathcal{S})$ at $\bm{\theta}^{(j-1)}$, $i.e.$, $\frac{\partial \mathcal{L}(\bm{\theta};\mathcal{S})}{\partial \bm{\theta}}|_{\bm{\theta}^{(j-1)}}$, are shown in Appendix~A.

\begin{algorithm}[t]
   \caption{Learning Time-varying Semi-parametric Hawkes Processes (TV-SPHP)}
   \label{alg1}
\begin{algorithmic}[1]
   \STATE \textbf{Input:} Event sequences $\mathcal{S}$. Weight $\gamma$. Learning rate $\delta$. Iteration number $J$.
   \STATE \textbf{Output:} Parameters $\bm{\theta}=[\bm{\mu}_{TI};\bm{\mu}_{TV};\bm{a}]$.
   \STATE Initialize $\bm{\theta}^{(0)}$ randomly from uniform distribution, $\bm{z}^{(0)}=\bm{\theta}^{(0)}$, $\bm{y}^{(0)}=\bm{0}$.
   \FOR{$j=1,...,J$}
   \STATE Gradient descent: $\bm{\theta}^{(j)}=(\bm{\theta}^{(j-1)}-\delta(\frac{\partial \mathcal{L}(\bm{\theta};\mathcal{S})}{\partial \bm{\theta}}|_{\bm{\theta}^{(j-1)}}+\rho(\bm{\theta}-\bm{z}^{(j-1)}+\bm{y}^{(j-1)})))_{+}$.
   \STATE $\min\frac{\rho}{2}\|\bm{\theta}^{(j)}+\bm{y}^{(j-1)}-\bm{z}\|_2^2+\gamma\|\bm{z}\|_1$ by soft-thresholding: $\bm{z}^{(j)}=S_{\gamma/\rho}(\bm{\theta}^{(j)}+\bm{y}^{(j-1)})$.
   \STATE Update dual variable: $\bm{y}^{(j)}=\bm{y}^{(j-1)}+\bm{\theta}^{(j)}-\bm{z}^{(j)}$.
   \ENDFOR
   \STATE $\bm{\theta}=\bm{\theta}^{(J)}$.
\end{algorithmic}
\end{algorithm}

\subsection{Simulation-based prediction}
Given the learned point process model and historical observations, we can predict the expected number of events in future time intervals.
In particular, for the target interval, we can simulate a set of synthetic event sequences based on Ogata's modified thinning method~\citep{ogata1981lewis}. 
The expected number of events in this interval can be estimated by the mean of the numbers of the events in these synthetic event sequences. 

It should be noted that when predicting future events based on our TV-SPHP model, we do not have access to the time-varying features for the future time interval.
Fortunately, for predictions in a short interval ($e.g.$, next 20 minutes or next one hour), we may assume that the time-varying features are relatively stable, allowing them to be approximated as the features corresponding to the last observed event. 
In summary, the scheme of our prediction method is shown in Algorithm~\ref{alg2}, in which the supremum of the intensity function in $[t, t_0+T]$ is approximated by discretization: $\sup_{s\in [t, t_0+T]}\lambda(s)\approx \max_{i=0,1,...,\frac{t_0+T-t}{\Delta}}\lambda(t+i\Delta)$, where the step size $\Delta$ is predefined.
\begin{algorithm}[t]
   \caption{Predicting the expected number of events in $[t_0, t_0+T]$}
   \label{alg2}
\begin{algorithmic}[1]
   \STATE \textbf{Input:} History $\mathcal{H}_{t_0}$. Learned model $\bm{\theta}$. Features $f_0$ and $f(t_0)$. Simulation trials $J$.
   \STATE \textbf{Output:} The expected number of events $\widehat{N}$.
   \STATE Initial $\widehat{N}=0$.
   \FOR{$j=1,...,J$}
   \STATE $t=t_0$.
   \REPEAT 
   \STATE Compute $m(t)=\sup_{s\in [t, t_0+T]}\lambda(s)$. Generate $s\sim \exp(m(t))$, $u\sim \mbox{Uniform}([0, 1])$.
   \STATE $t=t+s$. \textbf{If} $t<t_0+T$ \textbf{and} $u<\frac{\lambda(t)}{m(t)}$, \textbf{then} $\widehat{N}=\widehat{N}+1$.
   \UNTIL{$t>t_0+T$}
   \ENDFOR
   \STATE $\widehat{N}=\widehat{N}/J$.
\end{algorithmic}
\end{algorithm}

\begin{table}[t]
  \centering 
  \small{
  \caption{Prediction performance of all models}\label{tab2}
  \begin{tabular}{c|c|cccccc}\hline\hline
    \multirow{2}{*}{Method} & 
    \multirow{2}{*}{LogLike.} & 
    10 mins & 20 mins & 30 mins & 40 mins & 50 mins & 1 hr\\
    & & MAE & MAE & MAE & MAE & MAE & MAE \\
    \hline
    Poisson &-2132.7 &
    0.080 & 0.173 &
    0.259 & 0.381 &
    0.570 & 0.802\\
    Multi-task HP &-2093.9 &
    0.075 & 0.169 &
    0.246 & \textbf{0.356} &
    0.530 & 0.751\\
    TI-SPHP &-2095.6 &
    \textbf{0.074} & 0.174 &
    0.247 & 0.364 &
    0.545 & 0.763\\
    TV-SPHP &\textbf{-2058.4} &
    \textbf{0.074} & \textbf{0.167} &
    \textbf{0.243} & 0.365 &
    \textbf{0.523} & \textbf{0.733}
    \\ \hline\hline 
  \end{tabular}
  \label{tab:example} 
  }
\end{table}

\section{Experimental Results}

There were 42 subjects with valid data who jointly recorded a total of 5483 smoking events. Individual participants logged between 37 and 389 smoking events (1.9-19.5 packs), with a median of 113 (5.7 packs) and an interquartile range of 79 to 152 (4.0-7.6 packs).

The dataset was divided into $54$ smoking event sequences. 
Each sequence records the timestamps of a subject's smoking behaviors over a $4$ day period. 
The first $3$ days' events are used for training and the remaining events are used for testing. 
Based on different learned models, we predict the expected number of smoking events in the next time interval given the history of observations. 
The length of the time interval ranges from $10$ minutes to $1$ hour in $10$ minute increments, as shown in Table 2, to explore how the different models are affected. We feel that different windows in this range may be appropriate depending on the specific intervention (see Discussion).

To demonstrate the feasibility and superiority of our time-varying semi-parametric Hawkes process (\textbf{TV-SPHP}) method, we test it on our dataset and compare its performance with its variants and other state-of-the-art methods, including the parametric inhomogeneous Poisson process (\textbf{Poisson})~\citep{rathbun2016mixed}, the multi-task Hawkes process (\textbf{Multi-task HP})~\citep{xu2017benefits}, and the time-invariant semi-parametric Hawkes process (\textbf{TI-SPHP}). 
In particular, the Poisson method was designed for smoking event analysis, but only considers time-invariant and time-varying features while ignoring triggering patterns caused by previous smoking events. 
The multi-task HP does not utilize features and relies solely on temporal information, so that the exogenous part for each subject $n$ is represented as a value $\mu^n$ and learned directly. 
The TI-SPHP utilizes only the time-invariant features, $i.e.$ it ignores the term $\bm{\mu}_{TV}^{\top}\bm{f}^n(t)$ in (\ref{sphp}).

\begin{table}[t]
	\centering 
	\small{
		\caption{Standard deviation of prediction performance across simulation trials}\label{tab3}
		\begin{tabular}{c|cccccc}\hline\hline
			\multirow{2}{*}{Method} & 
			10 mins & 20 mins & 30 mins & 40 mins & 50 mins & 1 hr\\
			& SD & SD & SD & SD & SD & SD \\
			\hline
			Poisson &
			0.011 & 0.009 &
			0.010 & 0.017 &
			0.014 & 0.008\\
			Multi-task HP &
			0.007 & 0.008 &
			0.010 & 0.008 &
			0.010 & 0.013\\
			TI-SPHP &
			0.005 & 0.012 &
			0.011 & 0.011 &
			0.015 & 0.016\\
			TV-SPHP &
			0.005 & 0.005 &
			0.007 & 0.011 &
			0.012 & 0.011
			\\ \hline\hline 
		\end{tabular}
		\label{tab:example} 
	}
\end{table}

As previously mentioned, the time-invariant features include participant characteristics, which can be represented as $31$-dimensional ($C=31$) binary vectors. 
Each time-varying feature is an $8$-dimensional ($D=8$) vector consisting of a $4$-dimensional real-valued vector containing the participant's location (latitude and longitude), their speed, and the distance from their location to their home; and a $4$-dimensional binary vector representing the time of day.
The basis $\{\kappa_m\}$ are the Gaussian basis, with their number and bandwidth set using the method in~\citep{xu2016learning}.
In all of the above models, we apply an iterative optimization method to learn the parameters.
For our learning algorithm (Algorithm~\ref{alg1}), the number of iterations is $J=30$ and the learning rate is $\delta=0.01$.
The weight $\gamma=0.2$ is decided by cross-validation. 
For our prediction algorithm (Algorithm~\ref{alg2}), the number of simulation trials is $20$, the step size $\Delta$ is $10$ minutes.

\subsection{Model Performance}

For each testing sequence $k$, $k=1,...,K$, the true number of smoking events in a specific testing interval is $N_k$, and the expected number obtained from a certain learned model is $\widehat{N}_k$. Note that for the current dataset, $K$ is 54, the number of test sequences. We then calculate the following performance measures:
\begin{enumerate}
\item \textbf{MAE.} The mean absolute error of the estimated number:
\begin{eqnarray*}
\begin{aligned}
\mbox{MAE}=\frac{1}{K}\sideset{}{_{i=1}^{K}}\sum |N_k-\widehat{N}_k|.
\end{aligned}
\end{eqnarray*}
\item \textbf{Testing Log-Likelihood.} The log-likelihood values of the testing sequences are calculated for each model.
\end{enumerate}

Performance of all models is compared in Table 2. TV-SPHP performance was superior to the other models except for the 40-minute prediction window, and was superior to the Poisson model for all windows. Table 3 shows the standard deviation of the MAE across all simulation trials, demonstrating that results are consistent across testing sequences. In most cases, standard deviation is lowest for the TV-SPHP.

Binary classification accuracy (i.e. will smoke, will not smoke) was identical for all models -- values were 92.6\%, 85.2\%, 77.8\%, 64.8\%, and 53.7\% for 10-60 minute windows, respectively -- because smoking is a rare event with respect to these prediction windows, so all models heavily favor not smoking over smoking. For this reason, the MAE is the more appropriate measure of smoking risk in this application (see Discussion).

\subsection{Model Interpretation}

Model parameters are summarized and compared in Figure 1. Among the time-varying parameters, time of day was weighted much more heavily than the GPS measurements. The impact function, visualized in Figure 2, peaks within the first hour then decays over several hours, with marked daily periodicity.

\begin{figure}[h]
\centering
\includegraphics[width=.8\textwidth]{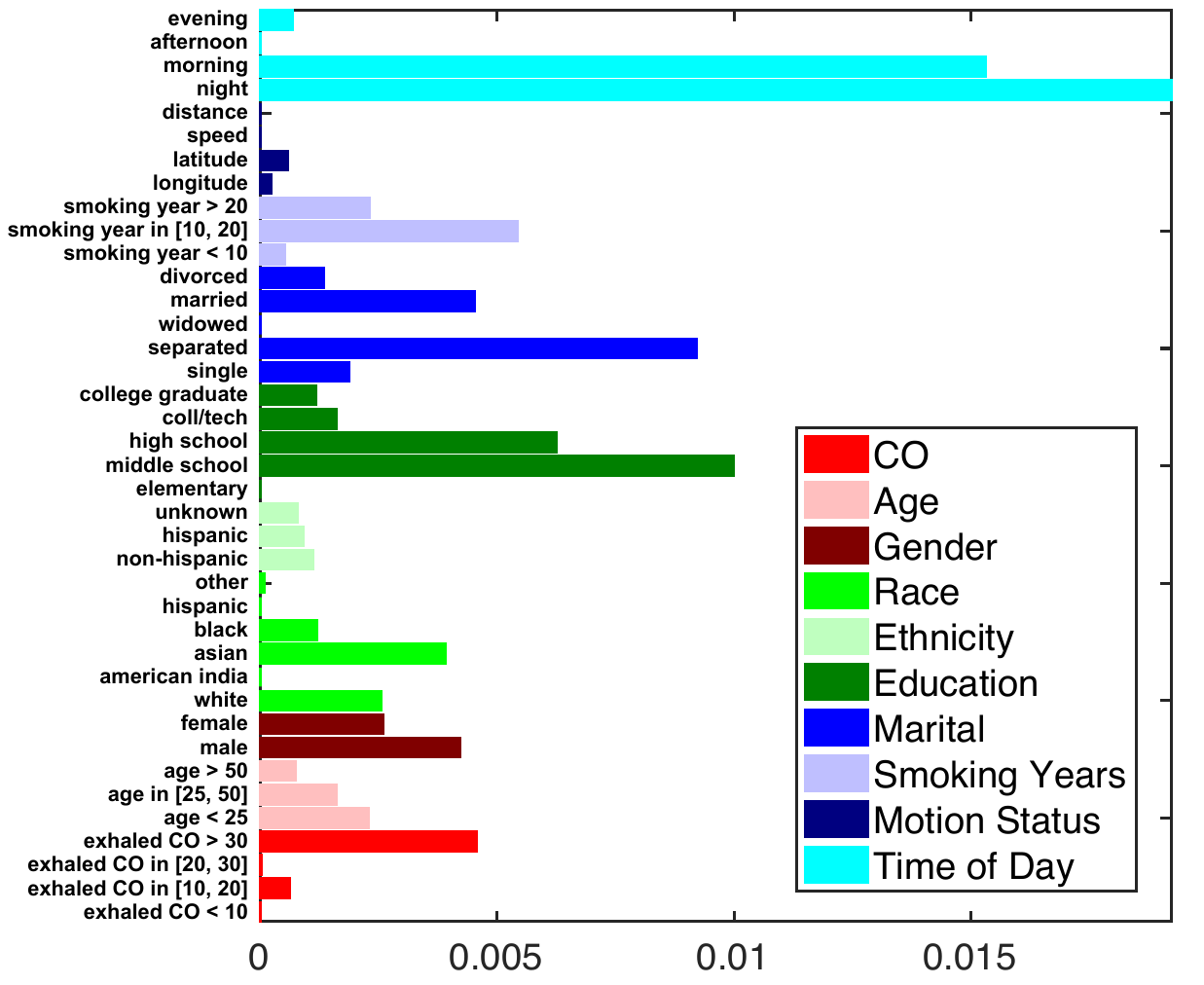}
\caption{Standardized coefficients for all exogenous factors}
\label{fig:coefficients}
\end{figure}

\begin{figure}[h]
\centering
\subfigure[Endogenous impact function]{
\includegraphics[width=.4\textwidth]{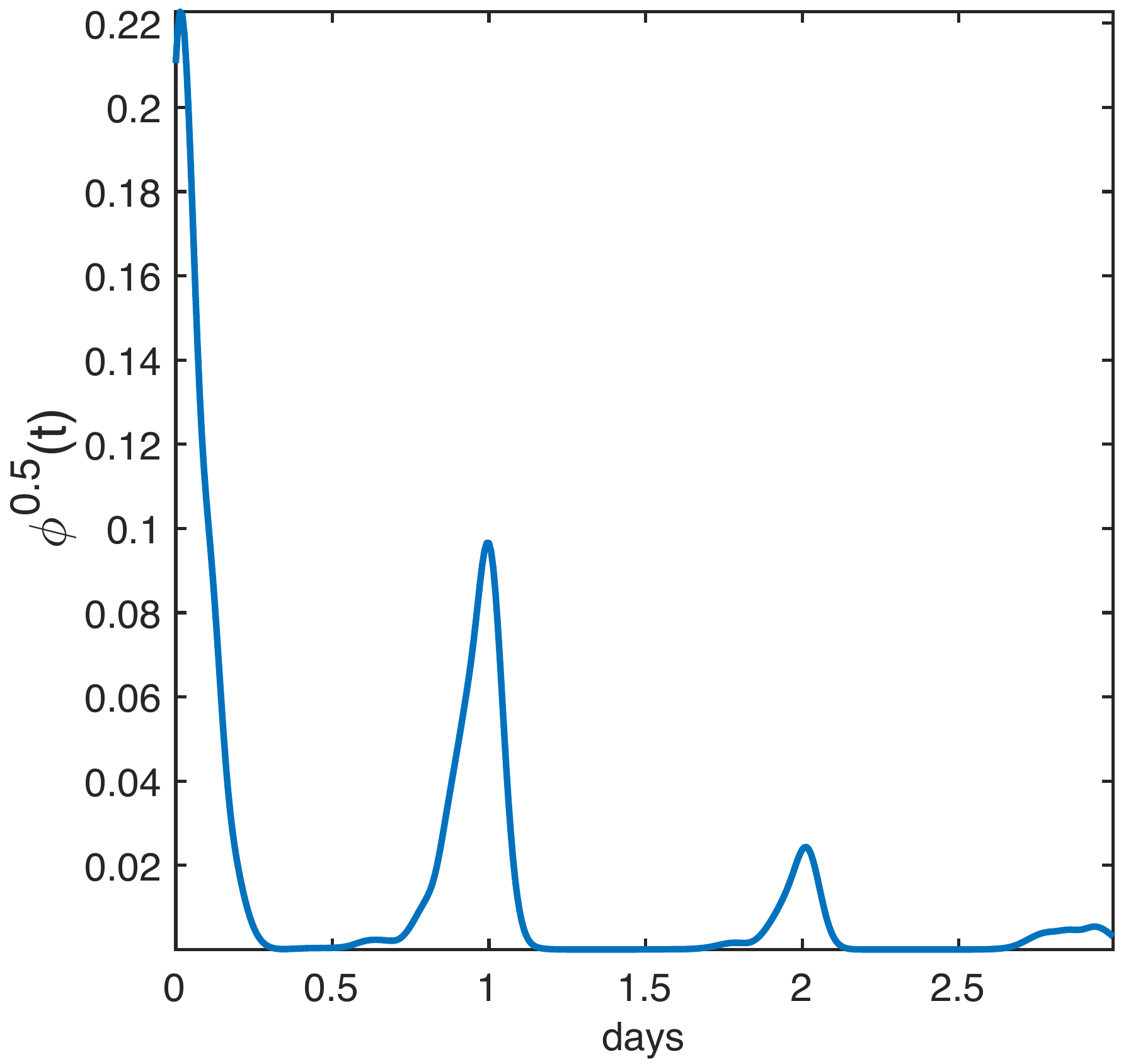} 
}
\subfigure[Enlarged endogenous impact function]{
\includegraphics[width=.4\textwidth]{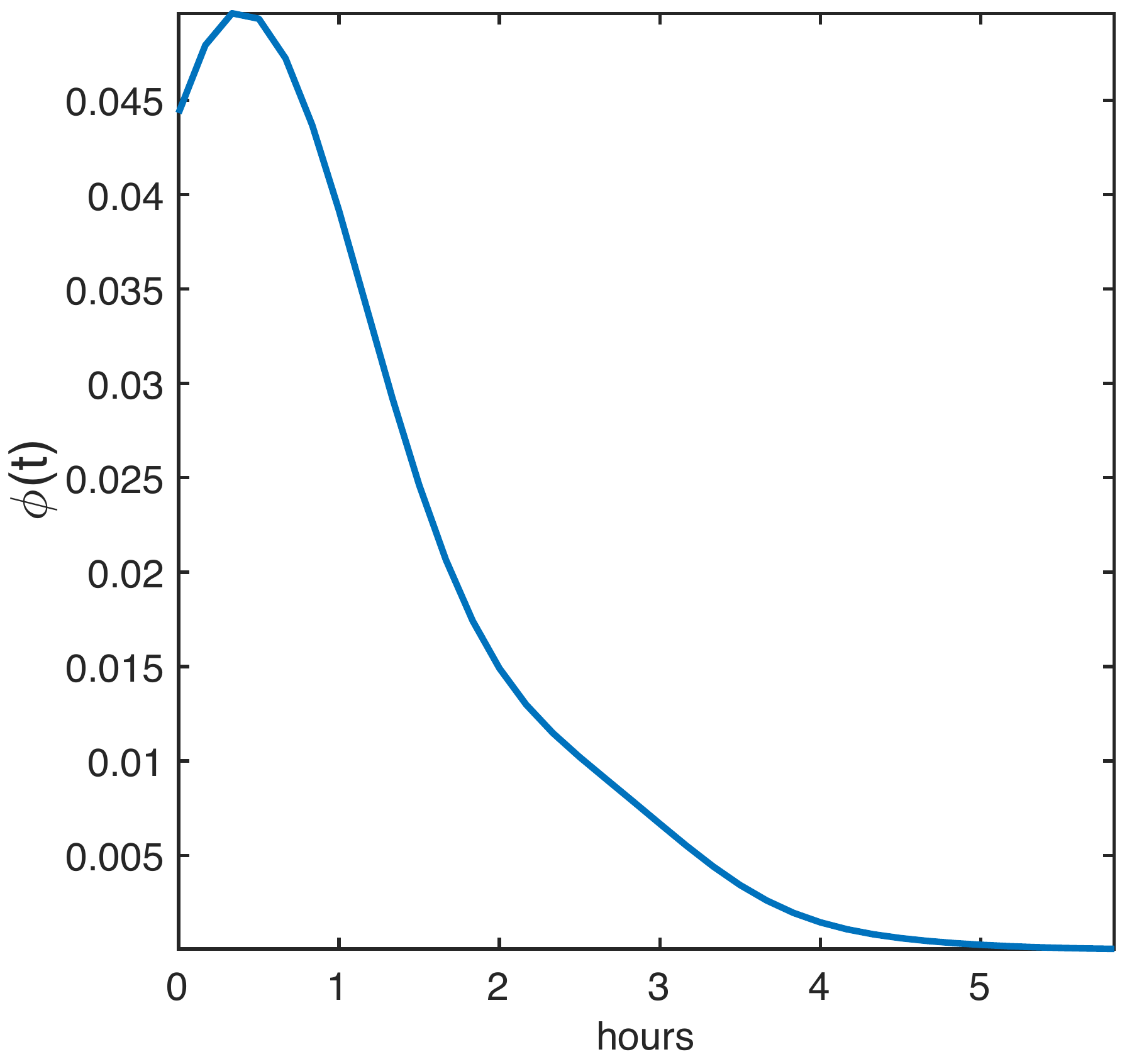} 
}
\caption{Visualization of (a) the learned impact function with length $3$ days, and (b) an enlarged view of the first $6$ hours. For better visual effect, we show the square root of the impact function, $i.e.$, $\phi^{0.5}(t)$, in the time domain.
}
\label{fig:impact} 
\end{figure}


\section{Discussion}




To our knowledge, this study is the first to model smoking as a self-triggering process, and among the first to evaluate the performance of smoking event prediction. Moreover, we have developed the time-varying semi-parametric Hawkes process (TV-SPHP), an extension of previous Hawkes process variants that is well-suited to this application. These methods and performance statistics can serve as baselines for future prediction models, leading to improved prediction of smoking risk. Effective prediction can be used to identify moments of high lapse risk and their causes/precursors, ultimately supporting more effective cessation strategies that could reduce death from smoking.

As hypothesized, prediction performance favored the TV-SPHP over its variants and previous methods, validating the advantages of modeling smoking as a self-triggering process. The models with and without time-varying features were similar over shorter prediction windows, but the TV-SPHP was superior over longer windows, most notably the 1 hour window. This suggests that time-varying features --- such as speed and time of day -- offer greater predictive advantage as the time since the last known smoking event increases.

The most appropriate prediction window depends on the application or intervention. A 10-minute window might fit a nicotine lozenge intervention due to its fast effect, whereas a longer window might be more appropriate for a self-help or peer support intervention. We feel that prediction windows longer than an hour are less reasonable, because (a) the time-varying features may change substantially during the window, and (b) smoking events occurring during the window affect the expected number of subsequent events.

Smoking is a rare event with respect to our prediction windows, therefore all models predict that the participant will not smoke in a given window. For this reason, accuracy (and other binary prediction measures) is a misleading measure of performance. The TV-SPHP most closely estimates the true number of cigarettes smoked (i.e. lowest MAE), which may be used as a measure of smoking risk. When placing a threshold on model output to trigger an intervention, a low threshold would be needed to achieve high sensitivity. This would also result in a high false positive rate, which may be acceptable given the importance of avoiding a lapse. The appropriate trade-off between these measures is intervention-specific.

The Hawkes process impact function is capable of capturing both short-term temporal dynamics, which may relate to nicotine metabolism and/or nicotine seeking, as well as longer-term daily patterns. For example, the peak $\sim$30 minutes after smoking may reflect nicotine seeking as plasma nicotine levels decline. This hypothesis could be explored in future work by testing how the learned impact function differs between groups with/without nicotine replacement. The location of this peak, which we hypothesized would occur within 1-4 hours of smoking, occurred earlier than expected. On average, plasma nicotine levels remain above $\frac{2}{3}$ of their peak value when the impact function peaks, but importantly, they are declining sharply.

The peaks 1 and 2 days after smoking, on the other hand, more likely reflect daily patterns and preferences rather than direct effects of nicotine. The clear daily periodicity of the impact function is striking, and corroborates previous observations that individual smokers exhibit consistent daily patterns of smoking \citep{Frederiksen1977,Robinson1980,Chandra2007}. The nature of the factors underlying these patterns (e.g. physiologic versus environmental) remains to be explored. 

\subsection{Limitations and Future Work}

A first and arguably most important limitation of this work is the uncertain reliability of self-reported smoking events. This problem is a well-known difficulty of detecting or predicting real-world smoking. Our pre-processing excluded participants with an unreasonable (high or low) number of button presses, but knowing the number is consistent with the participant's smoking habits does not guarantee that it is accurate. This step eliminated almost a third of participants (18/60), which underscores the difficulty of relying on self-report. Further, there is also no way to know whether the \textit{timing} of individual presses is accurate, which is particularly important given our focus on temporal dynamics. Research in smoking risk prediction will benefit from technologies that reliably detect smoking, bypassing the problems of self-report.

This pre-processing was essential to the goals of this study, but it also introduces a potential source of bias: participants with unpredictable smoking patterns may have been excluded because they appeared to be unreliable self-reporters. We have tried to mitigate this possibility by using a conservative exclusion threshold.

Finally, the measurements and participant populations in the parent studies were tailored to the objectives of those studies. We intend to extend this work with new data collection that includes a more comprehensive set of physiologic, self-report, and environmental measures. Many such factors are known to be associated with smoking risk and would be expected to improve prediction performance. A broader cohort of participants will be recruited, including those with significant health problems. These participants may be good candidates for intervention due to their more frequent interaction with the healthcare system.  Follow-up work will also explore more meaningful incorporation of location information via a hierarchical model that converts raw GPS coordinates to personal activity space (e.g. distinct locations visited by the participant throughout their day).

Future work will also explore relationships between nicotine metabolism (e.g. nicotine metabolite ratio) and the self-triggering nature of smoking, as quantified by the Hawkes process impact function. Nicotine metabolism is known to affect the relative efficacy of varenicline versus nicotine patch in smoking cessation \citep{Lerman2015}, for example. With our approach, the effect of pharmacologics and/or lapses on subsequent smoking risk could be explored through simulation, which may have important implications for cessation interventions.

\section{Conclusion}

In this work, we have developed the time-varying semi-parametric Hawkes process and applied it to the important problem of smoking prediction using data collected by 42 smokers. Motivated by the known influence of daily patterns and plasma nicotine levels on smoking behaviors, our approach models smoking as a self-triggering process. Results illustrate the temporal dynamics of smoking and demonstrate improved prediction performance compared to previous methods. Future work will extend the approach to a more comprehensive dataset and explore the effects of interventions such as nicotine replacement on temporal smoking dynamics. 

Effective smoking prediction models can pinpoint moments of high lapse risk in order to trigger a just-in-time intervention, for example with a mobile device. Modeling smoking as a self-triggering process may help to explain how a single lapse can progress to a failed quit attempt, thereby supporting cessation and reducing death from smoking.

 
\bibliography{sample,clinical_citations}

\section*{Appendix A.}
The gradient $\frac{\partial \mathcal{L}(\bm{\theta};\mathcal{S})}{\partial \bm{\theta}}|_{\bm{\theta}^{(j-1)}}$ is composed of $\frac{\partial\mathcal{L}}{\partial \mu_{TI,c}}$, $\frac{\partial\mathcal{L}}{\partial \mu_{TV,d}}$ and $\frac{\partial \mathcal{L}}{\partial a_m}$, where $\mu_{TI,c}$ and $\mu_{TV,d}$ are elements of $\bm{\mu}_{TI}$ and $\bm{\mu}_{TV}$, $c=1,...,C$, $d=1,...,D$, and $m=1,...,M$. 
In particular, in the $j$-th iteration we have
\begin{eqnarray*}
\begin{aligned}
&\frac{\partial\mathcal{L}}{\partial \mu_{TI,c}}=\sideset{}{_{n=1}^{N}}\sum\Bigl\{T f_{0,c}^n - \sideset{}{_{i=1}^{I_n}}\sum\frac{f_{0,c}}{\lambda^{n(j-1)}(t_i^n)}\Bigr\},\\
&\frac{\partial\mathcal{L}}{\partial \mu_{TV,d}}=\sideset{}{_{n=1}^{N}}\sum\sideset{}{_{i=1}^{I_n}}\sum\Bigl\{(t_i^n - t_{i-1}^n)f_{d}^n(t_i)-\frac{f_{d}(t_i^n)}{\lambda^{n(j-1)}(t_i^n)}\Bigr\},\\
&\frac{\partial\mathcal{L}}{\partial a_{m}}=\sideset{}{_{n=1}^{N}}\sum\sideset{}{_{i=1}^{I_n}}\sum\Bigl\{K_m(T-t_i^n)-\frac{\sum_{i'<i}\kappa_m(t_i^n-t_{i'}^n)}{\lambda^{n(j-1)}(t_i^n)}\Bigr\},
\end{aligned}
\end{eqnarray*}
where $\lambda^{n(j-1)}(t_i^n)$ is calculated based on (\ref{sphp}) given the parameters in previous iteration $\bm{\theta}^{(j-1)}$, $K^m(T-t_i^n)=\int_{0}^{T-t_i^n}\kappa_m(s)ds$, and $f_{0,c}^n$ and $f_{d}^n(t)$ are elements of $\bm{f}_0^n$ and $\bm{f}^n(t)$. 
Additionally, here we assume that the element of time-varying features are piecewise-constant functions over time, so we approximate $\int_{t_{i-1}^n}^{t_i^n}f_d^n(s)ds$ by $(t_i^n - t_{i-1}^n)f_{d}^n(t_i)$.

\end{document}